%% file: main.tex
\documentclass[letterpaper, 10 pt, conference]{ieeeconf}  % Comment this line out if you need a4paper

\IEEEoverridecommandlockouts                              % This command is only needed if 
\usepackage{cite}
\usepackage{amsmath,amssymb,amsfonts}
\usepackage{algorithmic}
\usepackage{graphicx}
\usepackage{textcomp}
\usepackage{xcolor}
\usepackage{balance}
\usepackage{comment}
\usepackage[ruled,vlined]{algorithm2e}
\usepackage{subcaption}
\usepackage{tensor}

\newcommand{\ie}{\emph{i.e.},}
\newcommand{\etal}{\emph{et~al.}}
\usepackage{hyperref}
\usepackage{float}
\usepackage{multirow}
\usepackage{colortbl}

\def\secref#1{Sec.~\ref{#1}}
\def\figref#1{Fig.~\ref{#1}}
\def\tabref#1{Table~\ref{#1}}
\def\eqref#1{Eq.~(\ref{#1})}

\definecolor{Gray}{gray}{0.85}
\definecolor{White}{gray}{1}

\newcolumntype{a}{>{\columncolor{Gray}}c}

\title{\LARGE \bf
Uncertainty-Aware Lidar Place Recognition in Novel Environments
}

\author{Keita Mason$^{1,2}$, Joshua Knights$^{1,2}$, Milad Ramezani$^{1}$, Peyman Moghadam$^{1,2}$, Dimity Miller$^{2}$ 
\thanks{$^{1}$Authors are with the Robotics and Autonomous Systems, DATA61, CSIRO, Brisbane, QLD 4069, Australia.
$^{2}$Authors are with the School of Electrical Engineering, Queensland University of Technology (QUT), Brisbane, Australia.} \thanks{Author contact emails: \tt{\footnotesize{ \{joshua.knights, milad.ramezani, peyman.moghadam\}\@@data61.csiro.au}}, \tt{\footnotesize{ \{kk.graves, d24.miller\}@qut.edu.au}}}%}
}

\begin{document}

\maketitle
\thispagestyle{empty}
\pagestyle{empty}

%%%%%%%%%%%%%%%%%%%%%%%%%%%%%%%%%%%%%%%%%%%%%%%%%%%%%%%%%%%%%%%%%%%%%%%%%%%%%%%%
\input{Sections/abstract.tex}
\input{Sections/introduction.tex}
\input{Sections/related_work.tex}
\input{Sections/methodology.tex}
\input{Sections/experimental_setup.tex}

\input{Sections/experiments.tex}
\input{Sections/conclusion.tex}

\section*{Acknowledgments}
The authors gratefully acknowledge funding of the project by the CSIRO's Machine Learning and Artificial Intelligence Future Science Platform. Dimity Miller acknowledges ongoing support from the QUT Centre for Robotics.

\balance{} 
\bibliographystyle{IEEEtran}
\bibliography{refs}

\end{document}

%% file: Sections/abstract.tex
\begin{abstract}
    State-of-the-art lidar place recognition models exhibit unreliable performance when tested on environments different from their training dataset, which limits their use in complex and evolving environments. To address this issue, we investigate the task of uncertainty-aware lidar place recognition, where each predicted place must have an associated uncertainty that can be used to identify and reject incorrect predictions. We introduce a novel evaluation protocol and present the first comprehensive benchmark for this task, testing across five uncertainty estimation techniques and three large-scale datasets. Our results show that an Ensembles approach is the highest performing technique, consistently improving the performance of lidar place recognition and uncertainty estimation in novel environments, though it incurs a computational cost. Code is publicly available at \href{https://github.com/csiro-robotics/Uncertainty-LPR}{https://github.com/csiro-robotics/Uncertainty-LPR}.
\end{abstract}

%% file: Sections/introduction.tex
\section{Introduction}

Localisation is a crucial capability of mobile robots – with an understanding of its location in a map, a robot can navigate to new locations, monitor an environment, and collaborate with other entities. Lidar place recognition (LPR) algorithms use point clouds to enable robot localisation – a robot can compare a recently captured point cloud with a map of previous point clouds to identify its current location, where the map can be generated on-the-fly or offline. These recognised locations,~\ie~revisit areas, can be used as loop closure constraints in a Simultaneous Localisation and Mapping (SLAM) algorithm to mitigate drift, or provide re-localisation in GPS-denied environments.

State-of-the-art approaches to LPR utilise deep neural networks~\cite{uy2018pointnetvlad, Zhang_2019_CVPR, Liu_2019_ICCV ,komorowski2021minkloc3d, vidanapathirana2021logg3d, transloc3d}, and exhibit impressive localisation performance when tested on environments included in the training dataset~\cite{komorowski2021minkloc3d, vidanapathirana2021logg3d, transloc3d}. However, when tested on a \emph{novel environment},~\ie~an environment not in the training dataset, their performance drops substantially. As an example, a MinkLoc3D model~\cite{komorowski2021minkloc3d} can achieve 93\% recall when trained and tested on urban road scenes from the United Kingdom, but this drops to 61\% recall when tested on urban road scenes from South Korea. This performance degradation highlights a critical weakness of state-of-the-art LPR techniques (and deep neural networks in general) -- an inability to generalise to conditions not represented in the training dataset.

\begin{figure}[t]
    \centering
    \includegraphics[width=1.0\linewidth]{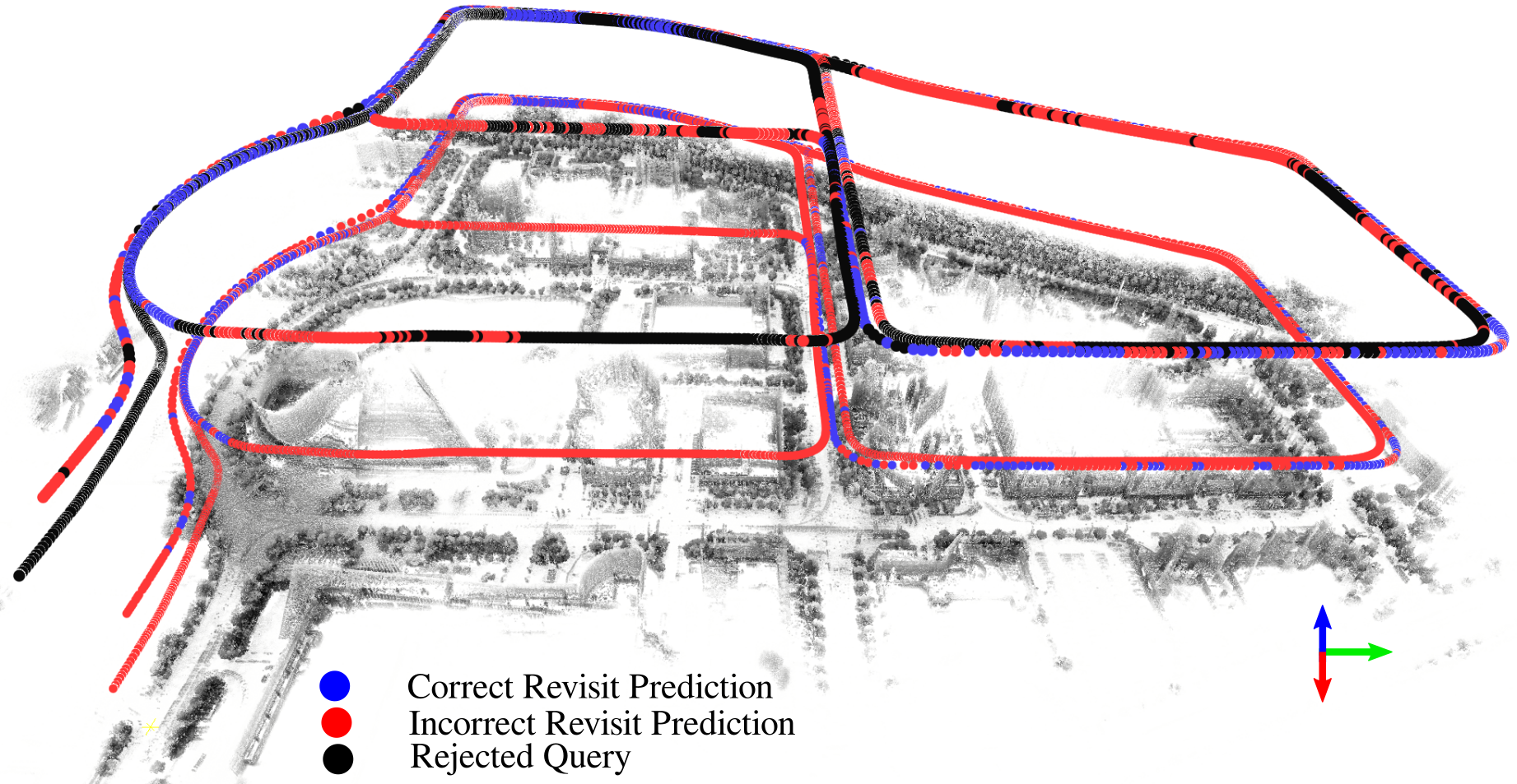}
    \caption{\small{The bottom traversal shows the performance of MinkLoc3D\cite{komorowski2021minkloc3d} when trained on Oxford road scenes, but tested on road scenes from the Daejeon Convention Centre in South Korea. Tested on this novel environment, MinkLoc3D predicts many false revisits. In the top traversal, we show uncertainty-aware LPR with Ensembles, where uncertainty is used to reject the 50\% most uncertain queries.}}
\label{fig:hero}
\vspace{-0.5cm}
\end{figure}

Uncertainty-aware deep networks are an established approach for enabling reliable performance in novel conditions~\cite{amodei2016concrete, sunderhauf2018limits, miller2018dropout}. Alongside each prediction, a network provides an estimate of its uncertainty, where high uncertainty indicates the network is more likely to make a mistake. While uncertainty-aware deep networks have been explored in many computer vision fields and robotics applications~\cite{lakshminarayanan2017simple, kendall2017uncertainties, miller2018dropout, miller2021uncertainty, shi2019probabilistic, cai2022stun,wirges2019capturing,cortinhal2020salsanext,meyer2020learning,pan2020towards}, no existing research explores uncertainty estimation for LPR. 

In this paper, we introduce uncertainty-aware lidar place recognition and present a comprehensive benchmark of this task. Our contributions are as follows: 
\begin{enumerate}
\item We formalise the uncertainty-aware LPR task for the prediction of lidar-based localisation failure (\secref{sec:method}).
\item Drawing inspiration from state-of-the-art techniques in related fields, we implement one standard baseline and four uncertainty-aware baselines for the LPR setting (\secref{sec:methodbaseline}).

\item We introduce an evaluation protocol that utilises three large-scale datasets and a range of metrics to quantify place recognition ability and uncertainty estimation for LPR in novel environments (\secref{sec:setup}).
\item We analyse the performance of all baseline methods on this evaluation protocol, exploring how different error types influence performance, as well as computational cost (\secref{sec:mainresults}).
\end{enumerate}

We hope that this work enables and stimulates further research into this important area by providing an extensive evaluation protocol and initial benchmark for comparison.

%% file: Sections/related_work.tex
\section{Related Work}
Before introducing uncertainty-aware LPR, we first contextualise the work by reviewing the existing state-of-the-art in LPR, and then discussing existing research into uncertainty estimation in retrieval tasks. 

\subsection{Lidar Place Recognition}
\label{related:pr}
LPR utilising 3D point clouds has been significantly explored in the last few years. LPR approaches identify similar places (revisited areas) by encoding high-dimensional point clouds into discriminative embeddings (often referred to as descriptors). Handcrafted LPR methods~\cite{salti2014shot, 6630945, 5152473, kim2018scan, gskim-2021-tro} generate local descriptors by segmenting point clouds into patches, or global descriptors that show the relationship between all the points in a point cloud. 

Recent state-of-the-art LPR approaches have been dominated by deep learning-based architectures due to their impressive performance~\cite{qi2017pointnet, Liu_2019_ICCV, Zhang_2019_CVPR, Hui2021PyramidPC, transloc3d, komorowski2021minkloc3d, komorowski2022improving, vidanapathirana2021logg3d, zhao2022spherevlad++, ma2022overlaptransformer}. These approaches typically utilise a backbone architecture to extract local features from the point cloud, which are then aggregated into a global descriptor. The specific design of these components varies significantly between different works; PointNet~\cite{qi2017pointnet}, graph neural networks~\cite{Liu_2019_ICCV}, transformers~\cite{Zhang_2019_CVPR,Hui2021PyramidPC, transloc3d}, and sparse-voxel convolutional networks~\cite{komorowski2021minkloc3d,komorowski2022improving,transloc3d,vidanapathirana2021logg3d} have all been proposed as local feature extractors, and aggregation methods include NetVLAD~\cite{Arandjelovi2018NetVLADCA}, Generalised Mean Pooling (GeM)~\cite{GeM2017} and second-order pooling~\cite{vidanapathirana2021locus,vidanapathirana2021logg3d}. 

\subsection{Uncertainty Estimation for Retrieval Tasks}
\label{related:ue}
Though there are a number of works exploring uncertainty estimation in lidar object detection~\cite{feng2019leveraging, feng2018towards, pan2020towards, wirges2019capturing, meyer2020learning} and point cloud segmentation~\cite{cortinhal2020salsanext}, no existing works explore uncertainty estimation for LPR. While recent work by Knights~\etal~\cite{knights2022incloud} shares a similar motivation to our work -- reliable performance in novel environments -- they explore incremental learning and specifically the issue of catastrophic forgetting.

Image retrieval is a field of computer vision that shares a similar problem setup to LPR (though notably operating on images rather than point clouds). 
When estimating uncertainty for image retrieval, recent works learn an uncertainty estimate by adding additional heads to their network architecture \cite{shi2019probabilistic, cai2022stun, warburg2021bayesian}. Shi~\etal~\cite{shi2019probabilistic} examine uncertainty-aware facial recognition, where face embeddings are modelled as Gaussian distributions by learning both a mean vector and variance vector. 

Warburg~\etal~\cite{warburg2021bayesian} follow a similar approach, introducing a `Bayesian Triplet Loss' to extend training to also include negative probabilistic embeddings. Most recently, STUN~\cite{cai2022stun} was proposed for uncertainty-aware visual place recognition. STUN presents a student-teacher paradigm to learn a mean vector and variance vector, using the average variance to represent uncertainty~\cite{cai2022stun}. Given the high performance of these approaches in the related image retrieval task, we adapt several of these methods to the LPR setting to serve as baselines for our benchmark.

%% file: Sections/methodology.tex
\section{Methodology}
\label{sec:method}
We first define the LPR task, and then formalise uncertainty-aware LPR. Following this, we introduce the baseline methods used for our benchmark.

\subsection{Lidar Place Recognition}
During LPR evaluation, a database contains point clouds with attached location information. This database can be a previously
curated map or can be collected online as an agent explores an environment. Given a query,~\ie~a new point cloud from an unknown location, an LPR model must localise the query by finding the matching point cloud in the database. If the predicted database location is within a minimum global distance to the true query location, the prediction is considered correctly \emph{recalled}. In this configuration, LPR performance is evaluated from the average recall of all tested queries~\cite{transloc3d, komorowski2021minkloc3d, uy2018pointnetvlad}.

We are motivated by the observation that perfect recall in an LPR setting does not currently exist, and may not be attainable in some applications -- when operating in dynamic and evolving environments, or dealing with sensor noise, the potential for error always exists. In this case, we argue that LPR models should additionally be able to estimate uncertainty in their predictions,~\ie~\emph{know when they don't know}. We formalise this below as uncertainty-aware LPR.

\subsection{Uncertainty-aware Lidar Place Recognition}
\label{sec:methodunc}
In uncertainty-aware LPR, each predicted match between a query and database entry should be accompanied by a scalar uncertainty estimate $U$.
This uncertainty represents the \emph{lack of confidence} in a predicted location.

Following the existing LPR setup, the primary goal in uncertainty-aware LPR is to maximise correct localisations (\ie~recall). Uncertainty-aware LPR extends on this by additionally requiring models to identify incorrect predictions by associating high uncertainty. We formulate this as a binary classification problem, where $U$ is compared to a decision threshold $\lambda$ to classify whether an LPR prediction is correct or incorrect:

\begin{equation}F_\lambda(U) = \left\{
\begin{array}{@{}ll@{}}
 \text{Correct}, & U \leq \lambda\\
 \text{Incorrect}, & U > \lambda\\
\end{array}
\right . \text{.}
\end{equation}  

Incorrect predictions can arise for two reasons: (1) the query is from a location that is not present in the database, or (2) the query is from a location in the database, but the LPR model selects the incorrect database match. We refer to these two error types as `no match error' and `incorrect match error' respectively, and analyse them in detail in \secref{sec:errortypesresults}.

\subsection{Baseline Approaches}
\label{sec:methodbaseline}
To benchmark uncertainty-aware LPR, we adapt a number of uncertainty estimation techniques existing in related fields to the LPR setting.

\noindent
\textbf{Standard LPR Network:} As explored in \secref{related:pr}, state-of-the-art LPR techniques utilise a deep neural network to reduce a point cloud to a descriptor $\mathbf{d}$. Given a database of $N$ previous point clouds and locations, a standard LPR network converts this to a database $\mathbb{D}$ of $N$ $L$-dimensional descriptors, $\mathbb{D} = \{\mathbf{d}_{i}\in \mathbb{R}^L\}_{i=1}^N$. During evaluation, a query point cloud  $\mathcal{P}_{q}\in \mathbb{R}^{M\times3}$, with $M$ points, is reduced to a query descriptor $\mathbf{d}_{q}\in \mathbb{R}^L$. This query descriptor is compared to the database descriptors, where the similarity between descriptors is measured via cosine similarity
\begin{equation}
\label{eq:sim}
    \mathrm{similarity}(\mathbf{d}_{q}, \mathbf{d}_{n}) = s_n= \frac{\mathbf{d}_{q} \cdot \mathbf{d}_{n}}{\|\mathbf{d}_{q}\| \|\mathbf{d}_{n}\|}, \  n\in\{1, ..., N\}.
\end{equation}

The database entry with the greatest similarity, $s_y$, is predicted as the matching location to the query. A naive measure of uncertainty from a standard LPR network is the negative cosine similarity between the query and the predicted database entry,
\begin{equation}
    U(\mathbf{d}_{q}, \mathbf{d}_{y}) = - s_{y}.
\end{equation}

\noindent
\textbf{Probabilistic Place Embeddings (PPE):} We implement Probabilistic Face Embeddings (PFE)~\cite{shi2019probabilistic}, an approach in the face recognition literature, for the LPR setting. PPE adds an additional uncertainty head to a pre-trained standard LPR network, using two fully-connected layers acting on the final layer of the network to output a descriptor uncertainty $\sigma$. When also treating the standard descriptor as a mean $\mu$, each descriptor can then be approximated as a Gaussian embedding, $\mathbf{d} = \mathcal{N}(\mu, \sigma)$. The added uncertainty head is trained on the dataset with a Mutual Likelihood Score loss, as described in \cite{shi2019probabilistic}.  

During testing, a Mutual Likelihood Score (MLS),~\ie~the log-likelihood of descriptors belonging to the same place, between the query descriptor and $n$-th database descriptor is calculated as 
\begin{equation}
\begin{aligned}
\label{eq:ppe}
    \mathrm{s_n} = &\log p(\mathbf{d}_{q} = \mathbf{d}_{n})\\
    = &-\frac{1}{2} \sum^L_{l = 1}(\frac{(\mu_q^{(l)} + \mu_n^{(l)})^2}{\sigma_q^{2(l)} + \sigma_n^{2(l)}} + \log(\sigma_q^{2(l)} + \sigma_n^{2(l)})) \\
    &- \frac{L}{2}\log 2 \pi.
\end{aligned}
\end{equation}

The database entry with the greatest MLS is predicted as the matching location to the queried point cloud. Uncertainty is measured as the negative MLS. 

\noindent
\textbf{Self-Teaching Uncertainty LPR Networks (STUN):} STUN was recently proposed for uncertainty-awareness in visual place recognition~\cite{cai2022stun}. We follow their approach for the LPR setting, using a pre-trained standard LPR network as the `Teacher', and creating a `Student' by adding an uncertainty head consisting of one fully-connected layer at the final layer of the original architecture.  Similar to the aforementioned PPE baseline, each descriptor is parameterised by an output mean $\mu$ and variance $\sigma$. The student network, with added uncertainty head, is trained with an uncertainty-aware loss, as per \cite{cai2022stun}. The average of the query descriptor's variance (over all dimensions), $U = \sum^L_{l=1}\sigma_q^{(l)}$, is used to represent uncertainty.

\noindent
\textbf{Dropout LPR Networks:} Monte-Carlo Dropout was introduced as a method for approximating a Bayesian Neural Network~\cite{gal2016dropout}, and has been adopted in a variety of computer vision fields due to its simplicity~\cite{miller2018dropout, kendall2017uncertainties, wirges2019capturing}. To use this technique for uncertainty-aware LPR, we train an LPR network with a dropout layer~\cite{srivastava2014dropout} placed in the backbone of the architecture. During evaluation, every point cloud is tested $M$ times to yield $M$ unique databases and $M$ unique queries. Building on \eqref{eq:sim}, this allows the computation of a set of $M \times N$ cosine similarity scores between the query descriptor and all $N$ database descriptors, 
\begin{equation}
   \{\{s_{n,m}\}^M_{m=1}\}^N_{n=1}.
\end{equation}

The cosine similarity score is averaged across the $M$ tests to produce the mean similarity between a query and database entry $n$,
\begin{equation}
    s_{n,\mu} = \frac{1}{M}\sum^M_{m=1}{s_{n, m}}.
\end{equation}

The database entry with the greatest mean similarity, $s_{y,\mu}$, is predicted as the matching location to the query and the negative mean similarity is used as the prediction's uncertainty.

\noindent
\textbf{Ensembles of LPR Networks:} Deep Ensembles~\cite{lakshminarayanan2017simple} was introduced by Lakshminarayanan~\etal~for uncertainty estimation in image classification and regression tasks. We extend it to uncertainty-aware LPR using a homogeneous ensemble of $M$ networks, with each network's weights initialised randomly, and the training dataset randomly shuffled during their training.

During evaluation, each member of the ensemble will produce a unique database and query descriptor, and thus we will have a unique set of $M$ similarity scores for all $N$ query-database comparisons. We approximate each query-database similarity as a Gaussian distribution, with mean and variance as 
\begin{equation}
\begin{aligned}
\label{eq:ens_mean}
    s_{n,\mu} &= \frac{1}{M}\sum^M_{m=1}{s_{n, m}},\\ 
    s_{n,\sigma^2} &= \frac{1}{M}\sum^M_{m=1}{(s_{n, m}-s_{n, \mu})^2}.
\end{aligned}
\end{equation}

The database entry with the greatest mean similarity, $s_{y,\mu}$, is predicted as the query location. The negative mean cosine similarity or similarity variance can be used to represent the uncertainty of this prediction. We find that the negative mean cosine distance performs best in practice, and thus use this when reporting results.

%% file: Sections/experimental_setup.tex
\section{Proposed Evaluation Protocol}
% \todo{Proposed Evaluation Protocol}
\label{sec:setup}
This section proposes a novel evaluation protocol designed to benchmark uncertainty-aware LPR in novel environments, detailing the datasets and metrics used, as well as experimental implementation specifics. 

\subsection{Datasets}
We utilise three publicly-available large-scale datasets for our evaluation -- the Oxford dataset~\cite{uy2018pointnetvlad}, the NUS Inhouse dataset~\cite{uy2018pointnetvlad}, and the MulRan dataset~\cite{Kim2020MulRanMR}. We detail the characteristics of each dataset below.

The \textbf{Oxford RobotCar dataset} is a subset of the Oxford RobotCar dataset~\cite{maddern20171} that was curated by Uy~\etal~\cite{uy2018pointnetvlad} for LPR. It contains point cloud submaps from 44 traversals through Oxford, U.K., collected over the course of a year. When evaluating performance on the Oxford RobotCar dataset, submaps from a single traversal are treated as queries and then evaluated iteratively with submaps from one of the remaining traversals as the database~\cite{uy2018pointnetvlad}. In this setup, every query has a database match and only `incorrect match' errors may occur. We follow the standard training and testing dataset split introduced by Uy~\etal~\cite{uy2018pointnetvlad}.

The \textbf{MulRan dataset}~\cite{Kim2020MulRanMR} consists of traversals of four different environments in South Korea -- the Daejeon Convention Center (\textbf{DCC}), the Daejeon city \textbf{Riverside}, the Korea Advanced Institute of Science and Technology (KAIST), and Sejong city (Sejong). We use MulRan to test in-session online LPR, where a traversal is evaluated sequentially and the database is collected over the course of the sequence. In this setup, not every query will have a database match and both `no match' and `incorrect match' errors may occur. We use the DCC and Riverside environments, training with DCC traversals 1 and 2 and testing on traversal 3, and training with Riverside traversals 1 and 3 and testing on traversal 2.\footnote{We do not test on the KAIST environment, as it can be considered ``solved'' with state-of-the-art techniques. Trained on any of the datasets (Oxford, DCC, or Riverside), MinkLoc3D~\cite{komorowski2021minkloc3d} can reliably achieve 97\% Recall@1 (or greater) when tested on KAIST. Additionally, we do not test on Sejong city, as it is not configured for in-session LPR.}

The \textbf{NUS Inhouse dataset}~\cite{uy2018pointnetvlad} consists of data from three different regions in Singapore -- a university sector, a residential area, and a business district. We use the standard refined test dataset split~\cite{uy2018pointnetvlad}. Point cloud submaps were collected across five traversals of each region at different times. Similar to evaluation on the Oxford dataset, submaps from a single traversal are used as queries and evaluated iteratively with submaps from one of the remaining traversals as the database~\cite{uy2018pointnetvlad}. Every query has a match and only `incorrect match' errors may occur.

Each dataset represents different urban areas and locations around the world. By training on one environment, and then testing performance on all other environments, we measure the performance of an algorithm on novel environments with various levels of similarity to the training environment. Specifically, we use three  different training environments -- (1) Oxford RobotCar, (2) DCC (MulRan), and (3) Riverside (MulRan), and four different evaluation environments -- (I) Oxford RobotCar, (II) DCC (MulRan), (III) Riverside (MulRan), (IV) NUS Inhouse. This provides a total of 12 possible training-evaluation dataset splits, where 9 test performance in novel environments. 

\subsection{Metrics}
As we formulated in \secref{sec:methodunc}, uncertainty-aware LPR has two performance criteria: (1) place recognition capability,~\ie~how accurately can a network find a query's matching entry in a database, and (2) uncertainty estimation,~\ie~does the network produce uncertainty that can be used to identify incorrect predictions. We detail below the three metrics we use to evaluate against these criteria.

\textbf{Average Recall@K} is an established metric for quantifying LPR performance~\cite{uy2018pointnetvlad, komorowski2021minkloc3d, transloc3d, knights2022incloud, vidanapathirana2021logg3d}. It calculates the portion of queries that are correctly localised, where at least one of the top-K database predictions is a ground-truth match with the query. Perfect performance is 100\%, and represents all possible revisits being correctly identified.

\textbf{Area under the Receiver Operating Characteristic curve (AuROC)} is a standard metric when evaluating the performance of uncertainty for binary classification~\cite{hendrycks2016baseline} -- given our formulation in \secref{sec:methodunc}, this makes it relevant for uncertainty-aware LPR. AuROC can be interpreted as the probability that an incorrect prediction has a greater uncertainty than a correct prediction~\cite{fawcett2006introduction}. It measures the area under the ROC curve, which plots the true positive rate (TPR) versus the false positive rate (FPR) across all possible uncertainty thresholds $\lambda$. Given a set of uncertainty estimates from correct revisit predictions ($\mathbb{U}_C$) and incorrect predictions ($\mathbb{U}_I$), the TPR and FPR can be calculated as

\begin{equation}
\label{eq:rec}
   TPR(\lambda) = \frac{|\mathbb{U}_C \leq \lambda|}{|\mathbb{U}_C|}
\end{equation}

\begin{equation}
\label{eq:prec}
   FPR(\lambda) = \frac{|\mathbb{U}_I \leq \lambda|}{|\mathbb{U}_C \leq \lambda| + |\mathbb{U}_I \leq \lambda|}.
\end{equation}
\noindent
Best performance is an AuROC of 100\%.

\textbf{Area under the Error vs Rejection curve (AuER)} measures both place recognition capability and uncertainty estimation. It is calculated from the area under the Error vs Rejection curve~\cite{grother2007performance}, also referred to a sparsification plot, which calculates the remaining error when progressively rejecting the most uncertain predictions. It can be constructed across a varying uncertainty-rejection threshold $\lambda$ by
\begin{equation}
   Error(\lambda) = \frac{|\mathbb{U}_I \leq \lambda|}{|\mathbb{U}_C \leq \lambda| + |\mathbb{U}_I \leq \lambda|}
\end{equation}

\begin{equation}
   Rejection(\lambda) = \frac{|\mathbb{U}_I > \lambda| + |\mathbb{U}_C > \lambda|}{|\mathbb{U}_C| + |\mathbb{U}_I|}.
\end{equation}

Fewer incorrect predictions will result in a lower error magnitude (thus measuring place recognition capability) and a well-calibrated uncertainty will reduce error while rejecting as few queries as possible (thus measuring uncertainty performance). Best performance is an AuER of 0.

\input{Sections/Results/main_table}

\subsection{Implementation Details}
We implement all baseline methods into the MinkLoc3D architecture~\cite{komorowski2021minkloc3d}, as it represents a lightweight state-of-the-art LPR architecture with strong performance over the range of datasets we use for benchmarking.  For our Ensembles and Dropout baseline, we use $M=5$ ensemble members and $M=5$ tests with dropout respectively.

For Dropout, we found that placing the dropout layer in MinkLoc3D between the bottom-up and top-down pass of the Feature Pyramid Network produced best results.

We follow the point cloud pre-processing procedures for Oxford RobotCar outlined in \cite{uy2018pointnetvlad,komorowski2021minkloc3d}, and pre-processing procedures for MulRan outlined in \cite{knights2022incloud}. When evaluating the Oxford and NUS Inhouse datasets, as per the established protocol in the literature, we define positive and negative training point clouds as within $10m$ and farther than $50m$ of the normalised centroid submaps, and a correct revisit prediction when a prediction is located within $25m$ of a positive ground truth~\cite{komorowski2021minkloc3d,uy2018pointnetvlad,transloc3d,knights2022incloud}. 
For the MulRan dataset, we follow \cite{knights2022incloud} and define positive and negative training point clouds within $10m$ and farther than $20m$ of the submap centroid. During evaluation, a query is compared against all point cloud submaps excluding the previous $90s$~\cite{knights2022incloud, vidanapathirana2021logg3d}, and we define a true positive revisit when a prediction is within $10m$ of the ground truth.

%% file: Sections/Results/main_table.tex
\setlength{\tabcolsep}{3.5pt}
\begin{table*}[t!]
\caption{\small{Across nearly all train-test dataset combinations, an Ensembles approach exhibits superior place recognition (measured by Recall@1) and uncertainty estimation (measured by AuROC and AuER). Best performance is indicated in bold.}}
\label{tab:comparison}
\begin{tabular}{clccccccccccccaaa}
\hline

\multicolumn{1}{c}{} & \multicolumn{1}{c}{} & \multicolumn{3}{c}{Oxford RobotCar} & \multicolumn{3}{c}{NUS Inhouse} & \multicolumn{3}{c}{DCC (MulRan)} & \multicolumn{3}{c}{Riverside (MulRan)} & \multicolumn{3}{c}{\cellcolor{Gray}Average} \\ 
 & \multicolumn{1}{c}{} & R@1 & AuROC & AuER & R@1 & AuROC & AuER & R@1 & AuROC & AuER & R@1 & AuROC & AuER & R@1 & AuROC & AuER \\
 \multicolumn{1}{c}{\textbf{Trained on:}} & \multicolumn{1}{c}{} & ($\uparrow$) & ($\uparrow$) & ($\downarrow$) & ($\uparrow$) & ($\uparrow$) & ($\downarrow$) & ($\uparrow$) & ($\uparrow$) & ($\downarrow$) & ($\uparrow$) & ($\uparrow$) & ($\downarrow$) & ($\uparrow$) & ($\uparrow$) & ($\downarrow$) \\\hline

  & Standard & 93.0 & 92.0 & 0.9 & 81.0 & 84.8 & 5.4 & 60.4 & 86.8 & 32.9 & 60.3 & 89.9 & 42.7 & 73.7 & 88.4 & 20.5 \\
\multirow{2}{*}{Oxford} & PPE & 92.4 & 90.9 & 1.1 & 78.1 & 83.9 & 6.8 & 61.6 & 87.9 & 31.0 & 60.3 & 91.4 & 41.5 & 73.1 & 88.5 & 20.1 \\
\multirow{2}{*}{RobotCar}& STUN & 91.4 & 90.5 & 1.3 & 75.9 & 84.1 & 7.8 & 62.9 & 87.9 & 30.0 & 60.3 & 89.5 & 42.2 & 72.6 & 88.0 & 20.3 \\
& Dropout & 92.2 & 90.5 & 1.1 & 81.4 & 85.1 & 5.1 & 60.5 & 86.2 & 33.2 & 61.0 & 89.4 & 42.6 & 73.8 & 87.8 & 20.5 \\
& Ensembles & \textbf{95.4} & \textbf{94.0} & \textbf{0.4} & \textbf{87.3} & \textbf{88.2} & \textbf{2.6} & \textbf{65.5} & \textbf{88.5} & \textbf{28.1} & \textbf{63.6} & \textbf{92.5} & \textbf{38.6} & \textbf{77.9} & \textbf{90.8} & \textbf{17.4} \\\hline
 
 \multirow{1}{*}{} & Standard & 64.8 & 82.1 & 14.1 & 75.9 & 85.7 & 7.4 & 86.0 & 96.7 & 13.4 & 65.6 & 93.8 & 36.9 & 73.1 & 89.6 & 17.9 \\
 \multirow{2}{*}{DCC} & PPE & 65.7 & 82.3 & 13.4 & 76.9 & 85.2 & 7.1 & 87.5 & 96.7 & 12.8 & 66.4 & 93.7 & 36.2 & 74.1 & 89.5 & 17.4 \\
\multirow{2}{*}{(MulRan)} & STUN & 65.7 & 82.7 & 13.1 & 77.0 & 85.4 & 7.0 & 87.3 & 96.7 & 12.9 & 66.1 & 94.4 & 35.7 & 74.0 & 89.8 & 17.2 \\
 & Dropout & 62.4 & 82.8 & 15.1 & 73.2 & 83.5 & 9.2 & 84.3 & 95.9 & 14.5 & 64.7 & 94.0 & 37.1 & 71.2 & 89.0 & 19.0\\
  & Ensembles & \textbf{70.1} & \textbf{85.4} & \textbf{9.7} & \textbf{81.7} & \textbf{87.6} & \textbf{4.7} & \textbf{88.0} & \textbf{97.2} & \textbf{12.4} & \textbf{66.6} & \textbf{94.9} & \textbf{35.3} & \textbf{76.6} & \textbf{91.3} & \textbf{15.5} \\
 \hline
 & Standard & 68.9 & 83.9 & 11.0 & 82.5 & 87.0 & 4.4 & 52.3 & 95.5 & 31.9 & 74.1 & 96.5 & 29.9 & 69.5 & 90.7 & 19.3 \\
 \multirow{2}{*}{Riverside} & PPE & 68.8 & 82.5 & 11.8 & 81.5 & 86.5 & 5.0 & 51.1 & \textbf{96.1} & 32.4 & 77.4 & 96.4 & 28.5 & 69.7 & 90.4 & 19.4 \\
\multirow{2}{*}{(MulRan)} & STUN & 68.1 & 83.7 & 11.5 & 80.7 & 86.2 & 5.4 & 50.8 & 95.9 & 32.7 & \textbf{78.7} & 96.0 & \textbf{28.1} & 69.6 & 90.5 & 19.4 \\
& Dropout & 66.9 & 81.4 & 13.2 & 82.7 & 86.9 & 4.4 & 52.1 & 95.5 & 32.1 & 75.3 & 95.6 & 30.0 & 69.3 & 89.9 & 19.9   \\
 & Ensembles & \textbf{73.5} & \textbf{86.9} & \textbf{7.7} & \textbf{86.3} & \textbf{89.0} & \textbf{2.8} & \textbf{53.7} & 95.2 & \textbf{31.0} & 74.9 & \textbf{97.0} & 29.2 & \textbf{72.1} & \textbf{92.0} & \textbf{17.7} \\
 \hline

\end{tabular}
\end{table*}

%% file: Sections/experiments.tex
\section{Experimental Results}
\label{sec:mainresults}

\begin{figure*}[t]
    \centering
    \includegraphics[width=0.85\linewidth]
    {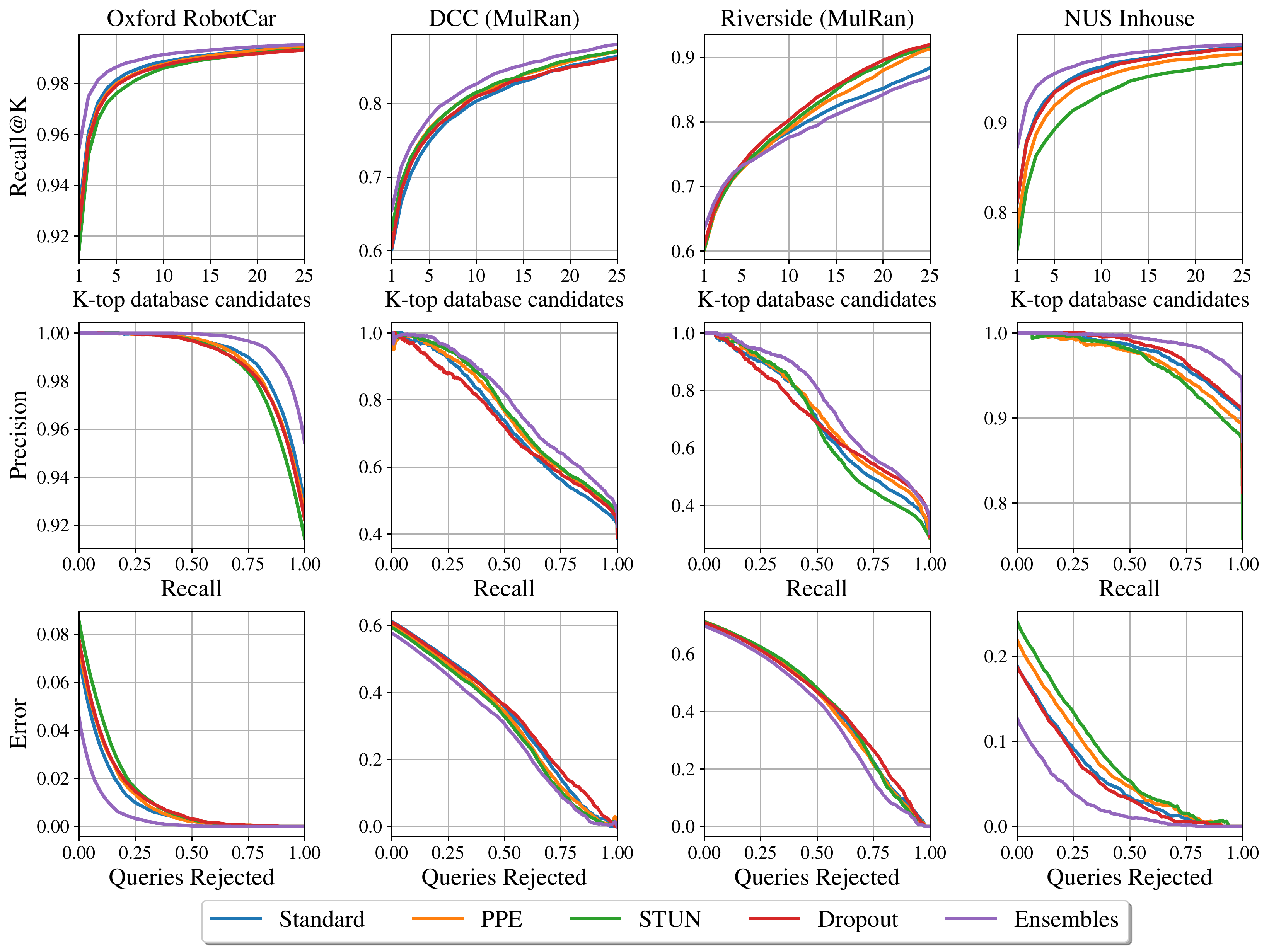}
    \caption{\small{Recall@K (top), precision-recall (middle) and error-vs-rejection (bottom) curves for the baselines when trained on Oxford RobotCar.}}
\label{fig:recallatk}
\end{figure*}

\subsection{Performance in Novel Environments}
\label{sec:result}
\tabref{tab:comparison} compares the performance of the uncertainty-aware LPR baselines across different train-test dataset configurations. We visualise Recall@K and error-versus-rejection curves for baselines trained on Oxford RobotCar in \figref{fig:recallatk}. We also show precision-recall curves~\cite{vidanapathirana2021logg3d, vidanapathirana2021locus} in \figref{fig:recallatk}. We highlight a number of trends from these results below.

\noindent
\textbf{Ensembles is the highest performing uncertainty-aware LPR baseline.} Across nearly all train-test data splits, Ensembles improve place recognition (Recall@1) and uncertainty estimation (AuROC and AuER) compared to all other baselines. On average, Ensembles has at least a 3\%, 1.7\% and 2\% absolute performance gain in Recall@1, AuROC, and AuER over all baselines. Considering the precision-recall curves in \figref{fig:recallatk}, Ensembles exhibit a greater precision across a range of recall values on all testing datasets.

\noindent
\textbf{PPE, STUN, and Dropout do not perform consistently across the different train-test configurations and metrics.} Depending on the train-test dataset combination, PPE, STUN and Dropout can increase, decrease, or not affect performance when compared to the Standard baseline -- across all three metrics. When training on DCC (MulRan), PPE and STUN both slightly improve Recall@1 by 0.9-1\% over the Standard baseline. On average, Dropout decrements Recall@1, AuROC, and AuER by 0.7\%, 0.7\%, and 0.9\% respectively when compared to the Standard baseline. We explore the behaviour of these baselines in more detail in \secref{sec:errortypesresults}, where we analyse uncertainty estimation for different error types.

\noindent
\textbf{Performance of all baseline methods decrements in novel environments.} When testing on Oxford RobotCar, DCC (MulRan) and Riverside (MulRan), all baseline methods exhibit decreased performance in place recognition and uncertainty estimation when training on a different environment. For example -- when training on Oxford RobotCar and testing on Oxford RobotCar, Ensembles exhibit 95.4\% Recall@1 and 94.0\% AuROC. However, when trained on DCC (MulRan) and testing on Oxford RobotCar as a novel environment, Ensembles performance decrements to 70.1\% Recall@1 and 85.4\% AuROC. This highlights the challenge of novel environments for uncertainty-aware LPR, and LPR in general, and motivates the need for future research to continue to bridge this gap.

\input{Sections/Results/errortypes}

\subsection{Performance on Different Error Types}
\label{sec:errortypesresults}
In \tabref{tab:errortypes}, we show the average results for DCC and Riverside (MulRan) when considering the two types of incorrect LPR predictions separately: (1) `incorrect match' errors, where a true match for the query exists in the database but was not predicted, and (2) `no match' errors, where no true match exists in the database. We report performance when training on DCC (MulRan), an environment similar to the test datasets, versus Oxford RobotCar, a novel environment. We highlight the key trends below.

% \noindent
\textbf{Ensembles improves uncertainty estimation for both incorrect match and no match errors.} Ensembles improves the Standard baseline uncertainty for incorrect match errors by 2.0-3.3\% AuROC and 1.5-4.8\% AuER, and by 0.2-1.1\% AuROC and 0.2-2.2\% AuER for no match errors.  

\textbf{STUN and PPE improve uncertainty estimation for incorrect match errors, but not for no match errors.} STUN consistently improves uncertainty estimation for incorrect match errors by 1.7-4\% AuROC and 1.3-3.3\% AuER. Tested in a novel environment (\ie~trained on Oxford RobotCar but tested on MulRan), PPE improves the incorrect match error performance by 2.9\% AuROC and 2.2\% AuER over the Standard baseline. In contrast, STUN and PPE either decrement performance or exhibit inconsistent performance improvements for uncertainty estimation for no match errors.

Notably, the Standard baseline exhibits already strong performance for no match errors -- between 93.6-98.9\% AuROC -- which may potentially offer uncertainty-aware techniques less room for improvement. In addition to this, different error types may be prone to different underlying uncertainty characteristics. For example, the technique of `learning' a variance to estimate uncertainty, as done by STUN and PPE, has been linked to aleatoric uncertainty~\cite{kendall2017uncertainties}. Aleatoric uncertainty describes uncertainty present in the input data, typically due to noise or randomness~\cite{kendall2017uncertainties}, which may be the underlying cause of most incorrect match errors. The other predominant type of uncertainty, epistemic uncertainty, describes uncertainty due to a lack of knowledge -- which intuitively may have greater relevance for no match errors. Ensembles estimates both aleatoric and epistemic uncertainty via predictive uncertainty~\cite{lakshminarayanan2017simple}, which may account for its high performance on both error types.

\textbf{Uncertainty-aware baselines offer greater improvements over a Standard baseline when operating in novel environments.} While we noted in \secref{sec:result} that all baselines reduced in performance when testing on novel environments, \tabref{tab:errortypes} shows that performance improvements of the uncertainty-aware baselines \emph{relative} to the Standard baseline are greater for novel environments. This suggests that while novel environments are the most challenging setting for uncertainty-aware LPR, it is also where state-of-the-art uncertainty estimation techniques may offer the greatest improvement. For example, Ensembles offer a greater uncertainty improvement when tested on novel environments (trained on Oxford RobotCar) by a factor of 1.7-5.5 for AuROC and 3.2-5.5 for AuER. Similarly, STUN improves uncertainty estimation in novel environments by a factor of 2.4 for AuROC and 3 for AuER for incorrect match errors. 

\subsection{Run-time for Uncertainty-aware LPR}

\input{Sections/Results/runtime}
\tabref{tab:runtime} quantifies the computational cost of each baseline, measured in the number of queries per second and milliseconds per query. As expected, the Standard baseline and STUN are the most computationally efficient, requiring only $3.1ms$ per query. PPE is only slightly slower, with $4.6ms$, due to its added calculation of the Mutual Likelihood Score between a query and all database descriptors. In contrast, Dropout and Ensembles are approximately linearly slowed by the number of models/query tests. Despite being the best-performing baseline, an Ensemble of 5 models requires $11.9ms$ per query when tested on an Nvidia Tesla P100-16GB GPU. 
For applications with limited computer availability, this may not allow real-time operation. 
This highlights a need for further research into uncertainty-aware LPR techniques that do not rely on sampling.

\subsection{Ensembles Ablation}
\label{sec:ensemble_ablation}
As the best-performing baseline, we now explore the performance of Ensembles in greater depth.

\textbf{Ensemble Size Impact on Performance:}
\begin{figure}[t]
    \centering
    \includegraphics[width=1.0\linewidth]
    {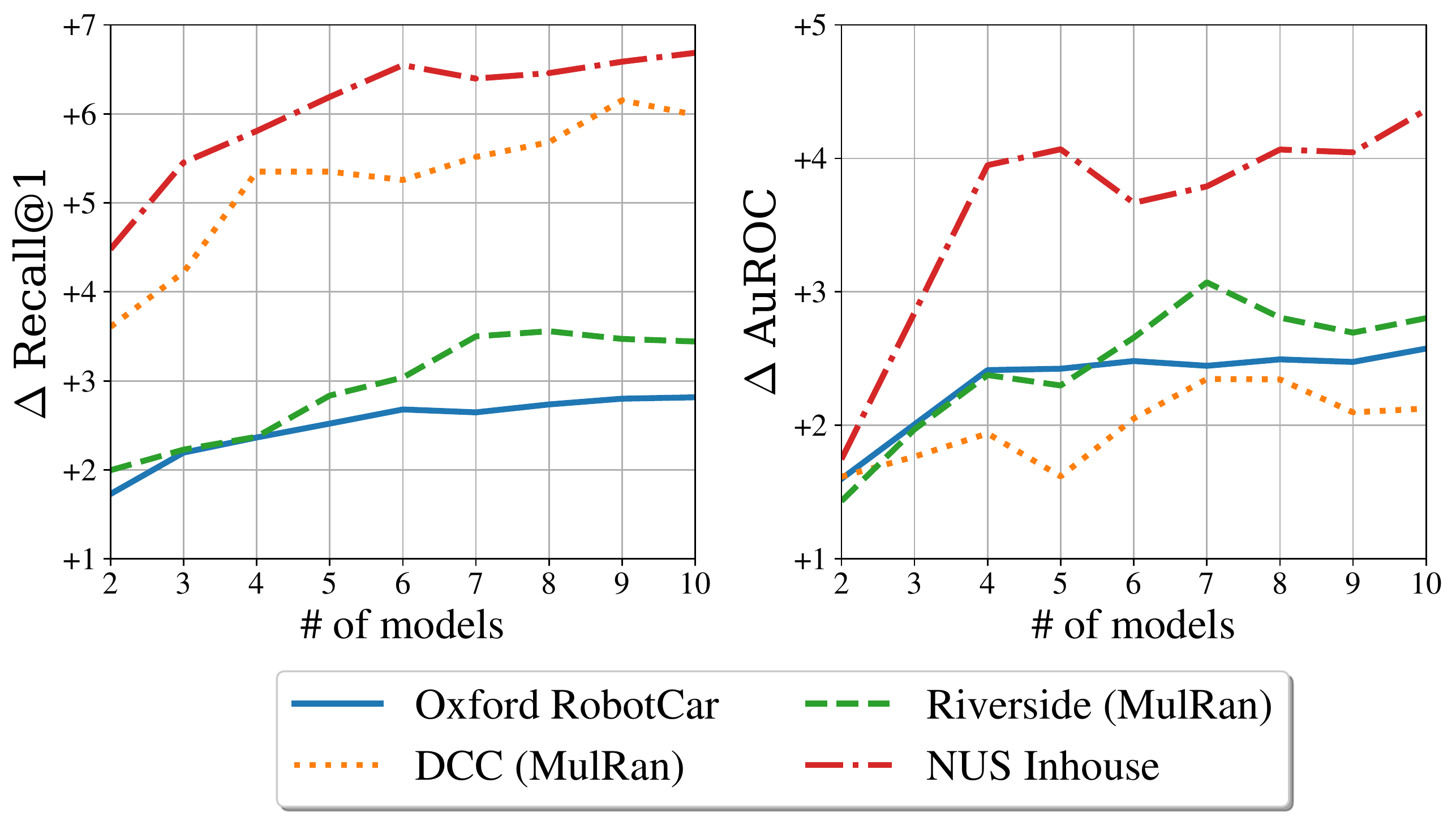}
    \caption{\small{Trained on Oxford RobotCar, we visualise how the Ensemble size influences the \emph{increase in performance} compared to the Standard baseline with a single model.}}
\label{fig:ablation}
\end{figure}

In \figref{fig:ablation}, we visualise the \emph{relative} performance increase of different Ensemble sizes when compared to the Standard baseline with a single model. Increasing the Ensemble size improves performance, although this performance change is not linear with increasing models. By 7 to 10 models, the performance increase slows or even plateaus for both place recognition capability (Recall@1) and uncertainy estimation (AuROC). Notably, even a 2-model Ensemble achieves an observable performance increase over a single model.

\textbf{Ensemble Performance with Different Architectures:}
\begin{figure}[t]
    \centering
    \includegraphics[width=1.0\linewidth]
    {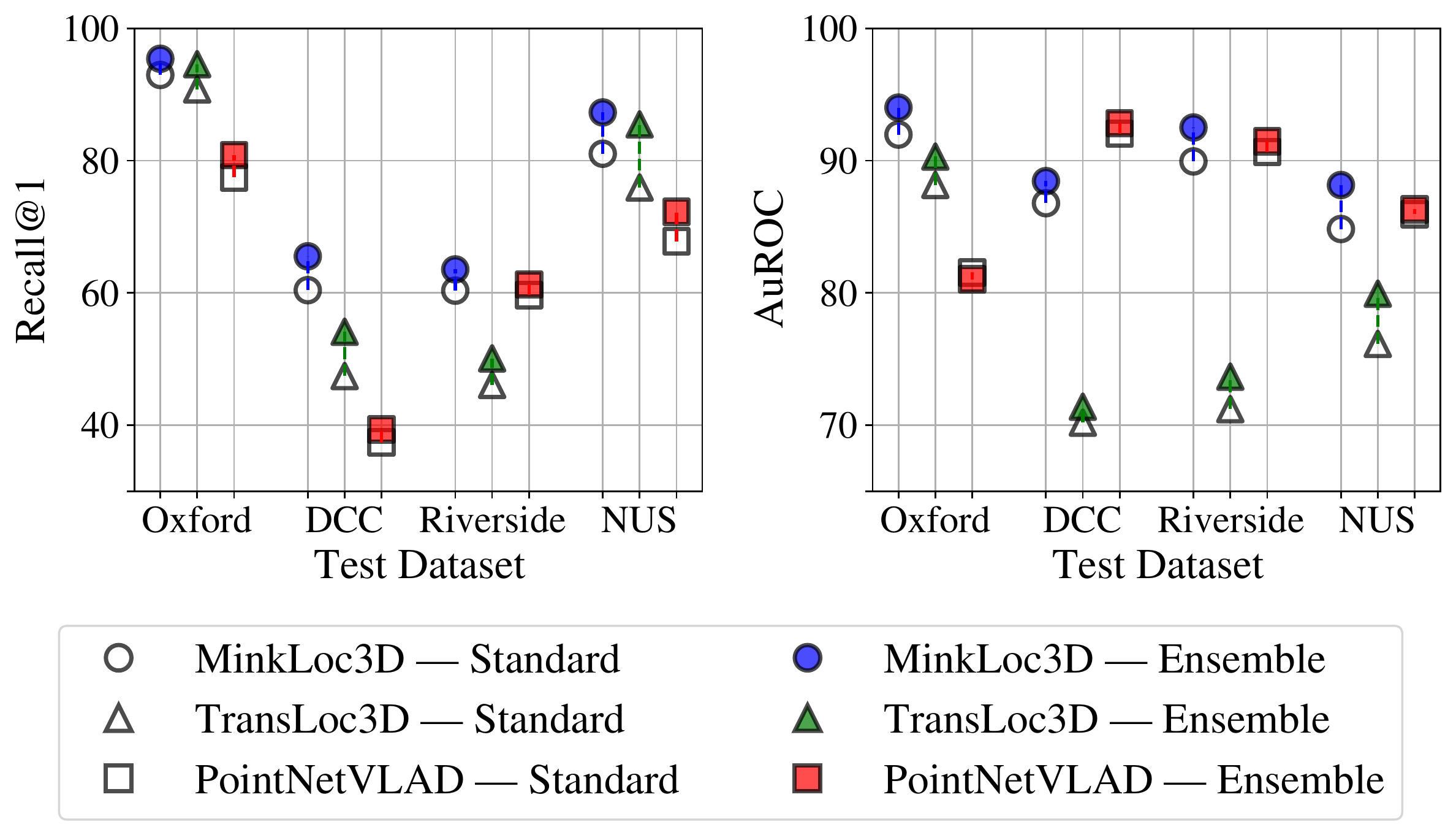}
    \caption{\small{For three different LPR architectures trained on Oxford RobotCar, we visualise the difference in Recall@1 and AuROC metrics for the Standard baseline versus Ensemble baseline.}}
\label{fig:architecture}
\end{figure}
Alongside MinkLoc3D~\cite{komorowski2021minkloc3d}, we implement the Standard and Ensembles baseline into two additional LPR architectures: one of the foundational LPR architectures, PointNetVLAD~\cite{uy2018pointnetvlad}, and a state-of-the-art transformer-based LPR architecture, TransLoc3D~\cite{transloc3d}. While \figref{fig:architecture} shows that an Ensembles approach improves Recall@1 and AuROC for all three architectures, the performance improvement is most notable for MinkLoc3D and TransLoc3D. We note that while TransLoc3D performs highly when the test dataset matches the training dataset (Oxford RobotCar), both place recognition capability and uncertainty estimation are significantly reduced for the test datasets representing novel environments (DCC, Riverside and NUS). Interestingly, while PointNetVLAD cannot compete with MinkLoc3D in place recognition capability (Recall@1), the Standard baseline exhibits very competitive performance in uncertainty estimation in novel environments (AUROC, on the DCC, Riverside and NUS Inhouse datasets). 

%% file: Sections/Results/errortypes.tex
\setlength{\tabcolsep}{3pt}
\begin{table}[t!]

\caption{\small{Performance of the uncertainty-aware baselines (relative to the Standard baseline) when categorised by error type.}}
\centering
\begin{tabular}{llcccc}
\hline
&           & \multicolumn{2}{c}{\textbf{Incorrect Match Error}} & \multicolumn{2}{c}{\textbf{No Match Error}} \\
\textbf{Trained on:}                              & \textbf{} & AuROC            & AuER            & AuROC         & AuER             \\
& & ($\uparrow$) & ($\downarrow$) & ($\uparrow$) & ($\downarrow$) \\\hline
\multicolumn{1}{c}{} & Standard  & 82.7              & 9.1               & 98.9              & 17.9             \\
\multicolumn{1}{c}{\multirow{2}{*}{DCC}}                              & $\Delta$PPE       & -0.2              & -0.6              & -0.2              & -0.2             \\
\multicolumn{1}{c}{\multirow{2}{*}{(MulRan)}}                              & $\Delta$STUN      & +1.7              & -1.3              & -0.3              & 0.0              \\
\multicolumn{1}{c}{}                              & $\Delta$Dropout   & +0.1              & +0.2              & -0.2              & +0.4             \\
\multicolumn{1}{c}{}                              & $\Delta$Ensembles & \textbf{+2.0}       & \textbf{-1.5}     & \textbf{+0.2}     & \textbf{-0.4}    \\ \hline
                  & Standard  & 78.6              & 19.4              & 93.6              & 24.5             \\
\multicolumn{1}{c}{\multirow{2}{*}{Oxford}} & $\Delta$PPE       & +2.9              & -2.2              & +0.4              & -0.5             \\
\multicolumn{1}{c}{\multirow{2}{*}{RobotCar}} & $\Delta$STUN      & \textbf{+4.0}     & -3.3              & -1.5              & +0.3             \\
& $\Delta$Dropout   & -3.5              & +1.1              & +0.5              & -0.5             \\
& $\Delta$Ensembles & +3.3              & \textbf{-4.8}     & \textbf{+1.1}     & \textbf{-2.2}    \\
\hline
\end{tabular}
\label{tab:errortypes}
\end{table}

%% file: Sections/Results/runtime.tex
\setlength{\tabcolsep}{5pt}
\begin{table}[t]
\centering
\caption{\small{Run-time of each baseline method when tested on a Nvidia Tesla P100-16GB GPU with batch size 1.}}
\label{tab:runtime}
\begin{tabular}{lrr}
\hline
Method  & Queries per second & Milliseconds per Query\\ \hline
Standard & 325 & 3.1\\
PPE & 219 & 4.6 \\
STUN & 325 & 3.1\\
Dropout\\
\multicolumn{1}{r}{5 tests} & 84 & 11.9\\
\multicolumn{1}{r}{10 tests}  & 43 & 23.3\\
Ensembles\\
\multicolumn{1}{r}{5 models}  & 84 & 11.9\\
\multicolumn{1}{r}{10 models}  & 44 & 22.7\\ \hline
\end{tabular}
\end{table}

%% file: Sections/conclusion.tex
\section{Conclusion}
\label{sec:conclusion}
In this paper, we formulated the task of uncertainty-aware lidar place recognition (LPR), where a network must also produce an uncertainty estimate to identify incorrect revisit predictions. We proposed a novel evaluation protocol that particularly focuses on performance in novel environments, and performed a comprehensive comparative analysis with five baseline methods adapted for the LPR task. Our results showed Ensembles consistently improves place recognition performance and uncertainty estimation, outperforming all other baselines. However, we also highlight that Ensembles are computationally expensive and may inhibit real-time operation in applications with limited compute availability. We hope that this benchmark motivates further research from the community that specifically targets the important problem of uncertainty-aware LPR. Future work could also explore how uncertainty-aware LPR can be used to infer uncertainty in robot pose estimation for loop-closure detection.